\documentclass{article}

\usepackage{mystyle}
 \usepackage[final]{nips_2018}

\usepackage[utf8]{inputenc} % allow utf-8 input
\usepackage[T1]{fontenc}    % use 8-bit T1 fonts
\usepackage{url}            % simple URL typesetting
\usepackage{booktabs}       % professional-quality tables
\usepackage{amsfonts}       % blackboard math symbols
\usepackage{nicefrac}       % compact symbols for 1/2, etc.
\usepackage{microtype}      % microtypography
\usepackage{mathtools}
\usepackage{subfigure}

\graphicspath{{fig/}}

\DeclareMathOperator{\mat}{mat}

\DeclareMathOperator{\col}{col}

\title{Provable Gaussian Embedding with One Observation}

\author{
  Ming Yu \thanks{Booth School of Business, University of Chicago, Chicago, IL. Email: \texttt{ming93@uchicago.edu} }\\
  \And
  Zhuoran Yang \thanks{Department of Operations Research and Financial Engineering, Princeton University, Princeton, NJ.}\\
  \And
  Tuo Zhao \thanks{School of Industrial and Systems Engineering, Georgia Institute of Technology, Atlanta, GA.} \\
  \And
  Mladen Kolar \thanks{Booth School of Business, University of Chicago, Chicago, IL.}\\
  \And
  Zhaoran Wang \thanks{Department of Industrial Engineering and Management Sciences, Northwestern University, Evanston, IL.} \\
}

\begin{document}

\maketitle

\begin{abstract}
  The success of machine learning methods heavily relies on having an appropriate representation for data at hand. Traditionally, machine learning approaches relied on user-defined heuristics to extract
  features encoding structural information about data. However,
  recently there has been a surge in approaches that learn how to
  encode the data automatically in a low dimensional
  space. Exponential family embedding provides a probabilistic
  framework for learning low-dimensional representation for various
  types of high-dimensional data \cite{rudolph2016exponential}. Though
  successful in practice, theoretical underpinnings for exponential
  family embeddings have not been established. In this paper, we study the Gaussian embedding model and develop the first theoretical results
  for exponential family embedding models. First, we show that, under
  mild condition, the embedding structure can be learned from
  \emph{one observation} by leveraging the parameter sharing between
  different contexts even though the data are \emph{dependent} with
  each other.  Second, we study properties of two algorithms used for
  learning the embedding structure and establish convergence results
  for each of them. The first algorithm is based on a convex
  relaxation, while the other solved the non-convex formulation of the
  problem directly. Experiments demonstrate the effectiveness of our
  approach.
\end{abstract}

\section{Introduction}
\label{sec:introduction}

Exponential family embedding is a powerful technique for learning a low
dimensional representation of high-dimensional data
\cite{rudolph2016exponential}. Exponential family embedding framework
comprises of a \emph{known} graph $G = (V, E)$ and the conditional exponential
family. The graph $G$ has $m$ vertices and with each vertex we observe
a $p$-dimensional vector $x_j$, $j=1,\ldots,m$, representing an
observation for which we would like to learn a low-dimensional
embedding.  The exponential family distribution is used to model the
conditional distribution of $x_j$ given the context
$\cbr{x_k, (k,j) \in E }$ specified by the neighborhood of the node
$j$ in the graph $G$. In order for the learning of the embedding to be
possible, one furthermore assumes how the parameters of the
conditional distributions are shared across different nodes in the
graph. The graph structure, conditional exponential family, and the
way parameters are shared across the nodes are modeling choices and
are application specific.

For example, in the context of word embeddings \cite{bengio2003neural,
  mikolov2013efficient}, a word in a document corresponds to a node in
a graph with the corresponding vector $x_j$ being a one-hot vector
(the indicator of this word); the context of the word $j$ is given by the
surrounding words and hence the neighbors of the node $j$ in the
graph are the nodes corresponding to those words; and the conditional
distribution of $x_j$ is a multivariate categorical distribution.
% The embedding vector then ``embeds'' each $x_j$ to a
% more interpretable low dimensional space. For example, it may contain
% ``gender'' as a component, and words ``man'', ``queen'' would have a
% large value in this component.
As another example arising in computational neuroscience consider
embedding activities of neurons. Here the graph representing the
context encodes spatial proximity of neurons and the Gaussian
distribution is used to model the distributions of a neuron's
activations given the activations of nearby neurons.

% where we can
% use Gaussian embedding to model the activities of neurons, and the
% contexts of a neuron are the nearby neurons and neurons that just
% activated. Moreover, in shopping data analysis, we can use Poisson
% embedding to model the purchase in each shopping cart, where the
% contexts of a specific item are the other items in the same shopping
% cart.

% The context of each node is defined as model specific and assumed to
% be \emph{known}. This known context structure forms a graphical model
% indicating the conditional dependency among nodes. The vector $x_j$ on
% node $j$ is conditional dependent with its context only, and the
% embedding vector embeds $x_j$ to a low dimensional vector space.

While exponential family embeddings have been successful in practice,
theoretical underpinnings have been lacking. This paper is a step
towards providing a rigorous understanding of exponential family
embedding in the case of Gaussian embedding. We view the framework of
exponential family embeddings through the lens of probabilistic
graphical models \cite{lauritzen1996graphical}, with the context graph
specifying the conditional independencies between nodes and the
conditional exponential family specifying the distribution locally.
%Our makes several contributions:
We make several contributions:
%\begin{itemize}

{\bf 1)} First, since the exponential family embedding specifies the
  distribution for each object conditionally on its context, there is
  no guarantee that there is a joint distribution that is consistent
  with all the conditional models. The probabilistic graphical models
  view allows us to provide conditions under which the conditional
  distributions defined a valid joint distribution over all the nodes.

{\bf 2)}  Second, the probabilistic graphical model view allows us to
  learn the embedding vector from one observation --- we get to see
  only one vector $x_j$ for each node $j \in V$ --- by exploiting the
  shared parameter representation between different nodes of the
  graph.  One might mistakenly then think that we in fact have $m$
  observations to learn the embedding. However, the difficulty lies in
  the fact that these observations are not independent and the
  dependence intricately depends on the graph structure. Apparently not
  every graph structure can be learned from one observation, however,
  here we provide sufficient conditions on the graph that allow us to
  learn Gaussian embedding from one observation.

{\bf 3)}  Finally, we develop two methods for learning the embedding. Our
  first algorithm is based on a convex optimization algorithm, while
  the second algorithm directly solves a non-convex optimization
  problem. They both provably recover the underlying embedding, but
  in practice, non-convex approach might lead to a faster algorithm.

%\end{itemize}

\subsection{Related Work}

\vspace{-1mm}
\paragraph{Exponential family embedding} Exponential family embedding
originates from word embedding, where words or phrases from the
vocabulary are mapped to embedding vectors \cite{bengio2003neural}.
Many variants and extensions of word embedding have been developed
since \cite{mikolov2013linguistic, levy2014neural, zou2013bilingual,
  levy2015improving}.  \cite{rudolph2016exponential} develop a probabilistic framework based on general exponential families that is
suitable for a variety of high-dimensional distributions, including
Gaussian, Poisson, and Bernoulli embedding. This generalizes the
embedding idea to a wider range of applications and types of data,
such as real-valued data, count data, and binary data
\cite{mukherjee2016gaussian, rudolph2017dynamic,
  rudolph2017structured}. In this paper, we contribute to the
literature by developing theoretical results on Gaussian embedding,
which complements existing empirical results in the literature.

\vspace{-1mm}
\paragraph{Graphical model.}

The exponential family embedding is
naturally related to the literature on probabilistic graphical models as
the context structure forms a conditional dependence graph among the
nodes. 
These two models are naturally related, but the goals and estimation procedures are very different. 
%\marginpar{add sentence "These..."}
Much of the research effort on graphical model focus on
learning the graph structure and hence the conditional dependency
among nodes \cite{lee2015learning, yang2015graphical,
  yu2016statistical, wang2016inference}. As a contrast, in this paper, we instead
focus on the problem where the graph structure is known and learn the
embedding.

\vspace{-1mm}
\paragraph{Low rank matrix estimation.}

As will see in Section \ref{sec:background}, the conditional
distribution in exponential family embedding takes the form
$f(VV^\top)$ for the embedding parameter $V \in \RR^{p \times r}$
which embeds the $p$ dimensional vector $x_j$ to $r$ dimensional
space. Hence this is a low rank matrix estimation problem.
%In low-rank matrix recovery literature people solve the following
%optimization problem
%\begin{equation}
%  \label{eq:opt:related}
%  \hat \Theta \in \arg \min_{\Theta \in \RR^{m_1 \times m_2}} f(\Theta)
%\quad \text{s.t. } {\rm rank}(\Theta) \leq r. 
%\end{equation}
%Although the objective function $f(\cdot)$ is usually (strongly)
%convex and smooth, this low-rank matrix recovery problem is non-convex
%and in general NP-hard \cite{fazel2004rank}. 
Traditional methods
focused on convex relaxation with nuclear norm regularization
\cite{negahban2011estimation, candes2009exact,
  recht2010guaranteed}. However, when the dimensionality is large,
solving convex relaxation problem is usually time consuming. 
%A more
%practical approach is to factorize $\Theta\in \RR^{m_1 \times m_2}$ as
%$\Theta = UV^\top$ where $U \in \RR^{m_1 \times r}$ and
%$V \in \RR^{m_2 \times r}$. This Burer-Monteiro type decomposition
%automatically imposes low-rank condition on $\Theta$.  
Recently there has been a lot of research on non-convex
optimization formulations, from both theoretical and empirical
perspectives \cite{wang2014nonconvex, Yu2017AnIM, sun2016guaranteed, yu2018learning, 
  zhao2015nonconvex}. People found that non-convex optimization is
computationally more tractable, while giving comparable or better
result.  In our paper we consider both convex relaxation and
non-convex optimization approaches.

\section{Background}
\label{sec:background}

\begin{figure}
    \centering
    \subfigure[Chain structure]
    {
        \includegraphics[width=0.3\textwidth]{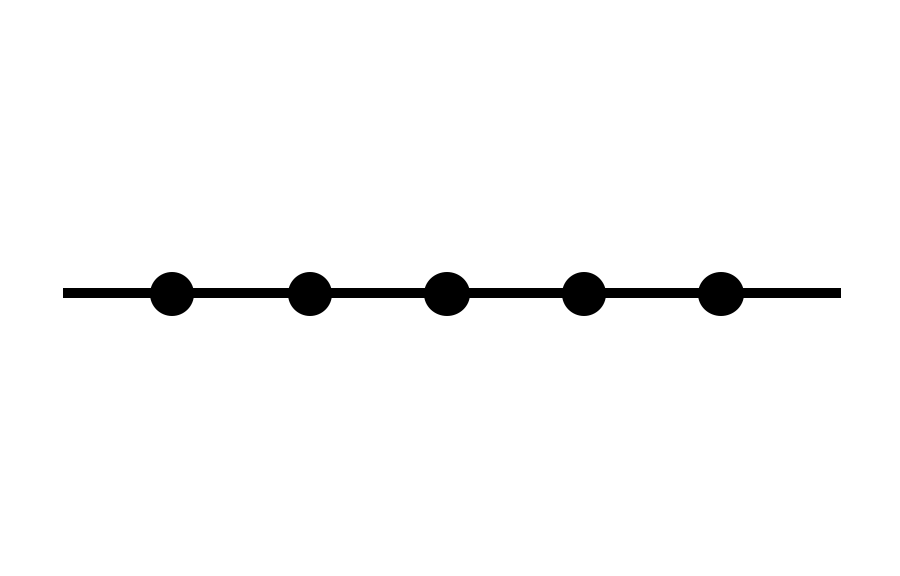}\hfill
        \label{fig:chain}
    }
    \,\,\,
    \subfigure[$\omega$-nearest neighbor structure]
    {
        \includegraphics[width=0.3\textwidth]{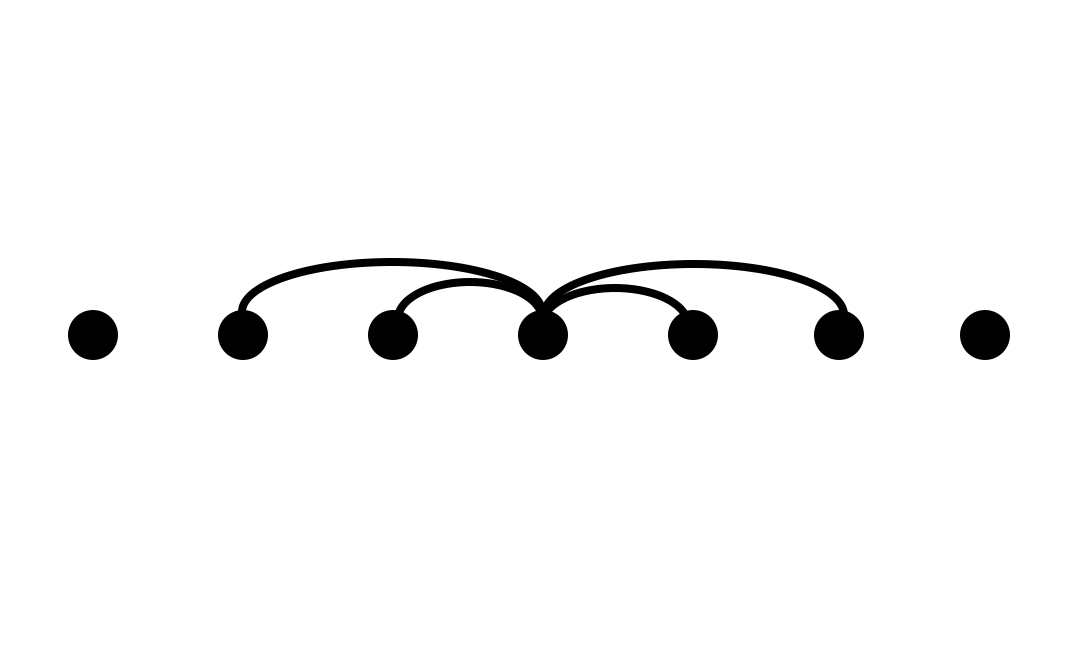}\hfill
        \label{fig:near}
    }
    \,\,\,
        \subfigure[Lattice structure]
    {
        \includegraphics[width=0.3\textwidth]{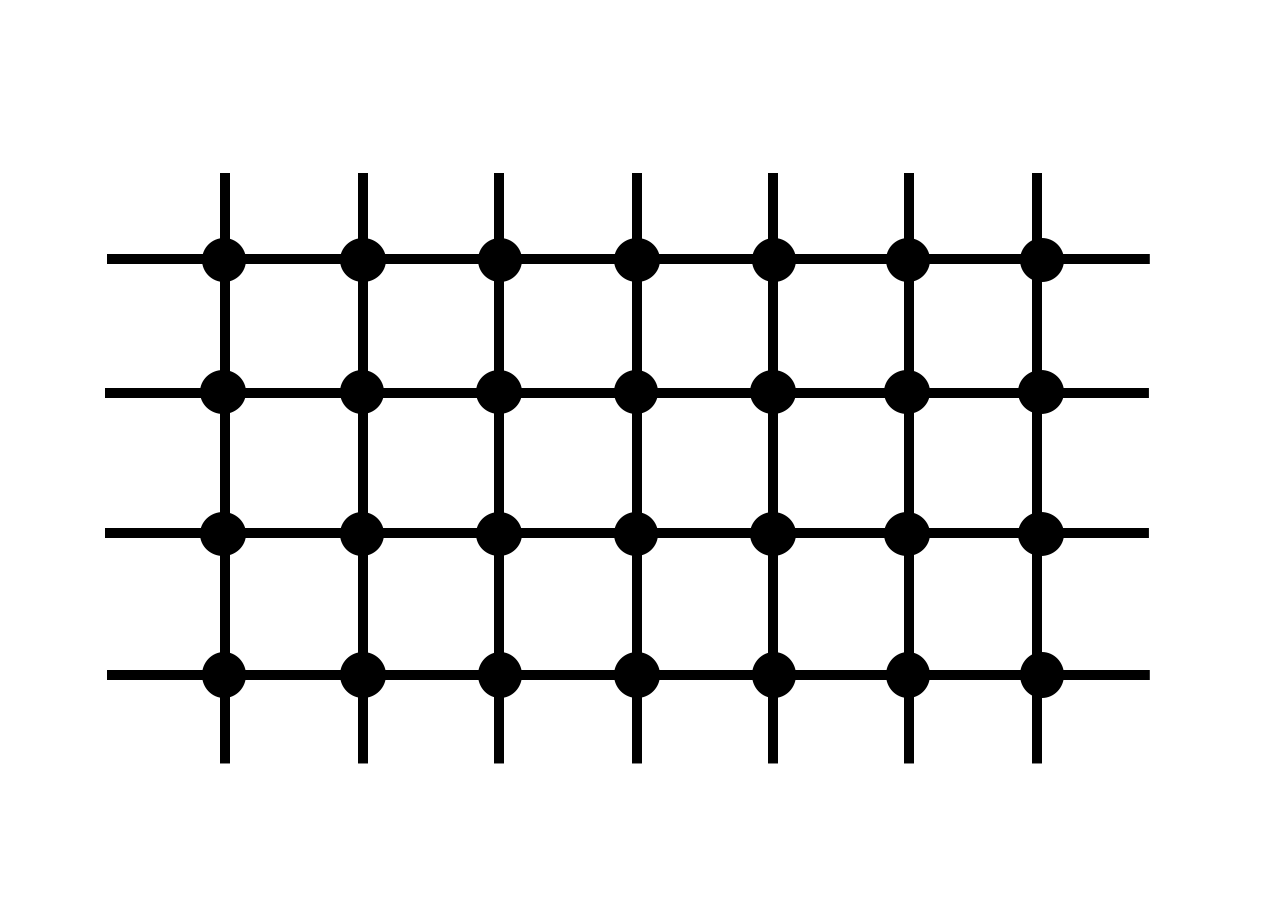}
        \label{fig:lattice}
    }
    \caption{Some commonly used context structures}
    \label{fig:context_structures}
     
\end{figure}

In this section, we briefly review the exponential family embedding
framework.  Let $ X = (x_1, \ldots, x_{m} ) \in \RR^{p \times m}$ be
the data matrix where a column $x_j \in \RR^p$ corresponds to a vector
observed at node $j$. For example, in word embedding, $x$ represents a
document consisting of $m$ words, $x_j$ is a one-hot vector
representation of the $j$-th word, and $p$ is the size of the
dictionary. For each $j$, let $c_j \subseteq \{1, ..., m\}$ be the
context of $j$, which is assumed to be known and is given by the graph
$G$ --- in particular, $c_j = \cbr{k \in V : (j,k)\in E}$.  
%\mcomment{Move figure to the top of the page. Do not call this figure 1,2,3 but figure 1 with a b and c panels.} 
Some commonly used context structures are shown in Figure \ref{fig:context_structures}.
%\ref{fig:chain} - \ref{fig:lattice}. 
Figure \ref{fig:chain} is for chain structure. Note
that this is different from vector autoregressive model where the
chain structure is directed.  Figure \ref{fig:near} is for
$\omega$-nearest neighbor structure, where each node is connected with
its preceding and subsequent $\omega$ nodes. This structure is common
in word embedding where the preceding and subsequent $\omega$ words
are the contexts. When $\omega = 1$ it boils down to the chain
structure. Finally Figure \ref{fig:lattice} is for lattice structure
that is widely used in the Ising model.

%\begin{subfigures}
%\begin{figure*}%[htbp]
%\begin{minipage}[t]{0.32\linewidth}
%\centering
%\includegraphics[width=0.75\textwidth]{chain}
%\caption{Chain structure}
%\label{fig:chain}
%\end{minipage}
%\begin{minipage}[t]{0.32\linewidth}
%\centering
%\includegraphics[width=0.75\textwidth]{near}
%\caption{$\omega$-nearest neighbor structure}
%\label{fig:near}
%\end{minipage}
%\begin{minipage}[t]{0.32\linewidth}
%\centering
%\includegraphics[width=0.75\textwidth]{lattice}
%\caption{Lattice structure}
%\label{fig:lattice}
%\end{minipage}
%\end{figure*}
%\end{subfigures}

The exponential family embedding model assumes that $x_j $
conditioning on $x_{c_j}$ follows an exponential family distribution
\vspace{-1mm}
\begin{equation}
\label{eq:model}
x_j \vert x_{c_j} \sim \texttt{ExponentialFamily} \Big[ \eta_j (x_{c_j}), t(x_j) \Big],
\end{equation}
where $t(x_j)$ is the sufficient statistics and
$\eta_j (x_{c_j}) \in \RR^p$ is the natural parameter. For the linear
embedding, we assume that $\eta$ in \eqref{eq:model} takes the form
\begin{equation}
\label{eq:model2}
\eta _j(x_{c_j}) = f_j \Big( V_j \sum_{k \in c_j}  V_k^\top x_k \Big ),
\end{equation}
where the link function $f_j$ is applied elementwise and
$V_{j} \in \RR^{p \times r}$.  The low dimensional matrix $V_k$ embeds
the vector $x_k \in \RR^p$ to a lower $r$-dimensional space with
$V_k^\top x_k \in \RR^r$ being the embedding of $x_k$.  For example,
in word embedding each row of $V_k$ is the embedding rule for a word.
Since $x_k$ is a one-hot vector, we see that $V_k^\top x_k $ is
selecting a row of $V_k$ that corresponds to the word on the node $k$.
A common simplifying assumption is that the embedding structure is
shared across the nodes by assuming that $V_j = V$ for all $j\in V$.
In word-embedding, this makes the embedding rule not depend on the
position of the word in the document.
We summarize some commonly seen exponential family distributions and
show how they define an exponential family embedding model.

\paragraph{Gaussian embedding.} 
In Gaussian embedding it is assumed that the conditional distribution is

\#
\label{eq:conditional_Gaussian}
 x_j |  x_{c_j} \sim N\Big(V\sum_{k\in c_j} V^\top x_k,  \Sigma_j\Big) = N\Big( M\sum_{k\in c_j}  x_k,  \Sigma_j\Big),
\# 
where $ M = VV^\top$ and $ \Sigma_j$ is the conditional covariance
matrix for each node $j$. We will prove in Section~\ref{sec:model}
that under mild conditions, these conditional distributions define a
valid joint Gaussian distribution. The link function for Gaussian
embedding is the identity function, but one may choose the link function to be
$f(\cdot ) = \log (\cdot)$ in order to constrain the parameters to be
non-negative. Gaussian embedding is commonly applied to real valued observations.

\paragraph{Word embedding (cbow \cite{mikolov2013efficient}).}

In the word embedding setting, $x_j$ is an indicator of the $j$-th
word in a document and the dimension of $x_j$ is equal to the size of the vocabulary.
The context of the $j$-th word, $c_j$, is the window of size $\omega$
around $x_j$, that is, $c_j = \{k\in \{1, ..., m\} \colon k \neq j, | k - j | \leq \omega \}$.
Cbow is a special case of exponential family embedding with
\begin{equation}
\begin{small}
\label{eq:cbow} p(x_j \vert x_{c_j} ) = \frac{ \exp \left [ x_j ^\top V \left ( \sum_{k\in c_j } V^\top x_k \right ) \right ] }{ \sum_{j } \exp \left [ x_j ^\top V \left ( \sum_{k\in c_j } V^\top x_ k \right ) \right ] }.
\end{small}
\end{equation}

\paragraph{Poisson embedding.} 
%\mcomment{Write down distribution form like in the case of two previous embeddings. }
In Poisson embedding, the sufficient 
statistic is the identity and the natural parameter is the logarithm
of the rate. The conditional distribution is given as
\vspace{1mm}
\#
 x_j |  x_{c_j} \sim \text{Poisson}\Big(\exp \big(V\sum_{k\in c_j} V^\top x_k \big)\Big).
\#
% There are two choices of the link function: identity and $\log$.
%Similar as Gaussian embedding, by taking the link function $f(\cdot ) = \log (\cdot)$ we get nonnegative Poisson embedding, where we only capture positive correlations between nodes. 
%Logarithmic link correspond to embeddings with nonnegative parameters.

\vspace{-3mm}
Poisson embedding can be applied to count data.

\section{Gaussian Embedding Model}
\label{sec:model}

In this paper, we consider the case of Gaussian embedding, where the
conditional distribution of $x_j$ given its context $x_{c_j}$ is given
in \eqref{eq:conditional_Gaussian} with the conditional covariance
matrix $\Sigma_j$ unknown. The parameter matrix $M = VV^\top$ with
$V \in \RR^{p \times r}$ will be learned from the data matrix $X \in \RR^{p \times m}$ and
$V^\top x_k$ is the embedding of $x_k$.

Let
$X_{\col} = [ x_1^\top, x_2^\top, ..., x_m^\top]^\top \in \RR^{pm
  \times 1}$ be the column vector obtained by stacking columns of $X$
and let $[ x_j]_{\ell}$ denote the $\ell$-th coordinate of $x_j$.
We first restate a definition on compatibility from \cite{wang2008conditionally}.

\begin{definition}
A non-negative function $g$ is capable of generating a conditional density function $p(y | x)$ if
\vspace{-1mm}
\#
p(y|x)= \frac{g(y,x)}{\int g(y, x)dy}.
\#
Two conditional densities are said to be compatible if there exists a function $g$ that can generate both conditional densities. When $g$ is a density, the conditional densities are called strongly compatible.
\end{definition}
%From the conditional distribution of $ x_j \in \RR^p$ given in \eqref{eq:conditional_Gaussian},
%we observe that the conditional distribution $[ x_j]_{\ell}$ given all the other components of
%$X_{\col}$ is 
%\begin{equation}
%\begin{aligned}
%  \label{eq:conditional_Gaussian_full}
%  [ x_j]_\ell \, \Big| \, x_{c_j}, [ x_j]_{\ell^c} \sim N\bigg( \Big[ M\sum_{k\in c_j}  x_k\Big]_\ell &+ [ \Sigma_j]_{\ell,\ell^c}[ \Sigma_j]_{\ell^c,\ell^c}^{-1} \cdot \Big([ x_j]_{\ell^c} - \Big[ M\sum_{k\in c_j}  x_k\Big]_{\ell^c}\Big),  \\
%  &[ \Sigma_j]_{\ell,\ell} - [ \Sigma_j]_{\ell,\ell^c}[
%  \Sigma_j]_{\ell^c,\ell^c}^{-1}[ \Sigma_j]_{\ell^c,\ell}\bigg),
%\end{aligned}
%\end{equation}
Since $M$ is symmetric, according to Proposition 1 in \cite{chen2014selection}, we have the following proposition. 

%\mcomment{This proposition somehow does not make sense.}
\begin{proposition}
  \label{lemma:compatible}  
  The conditional distributions \eqref{eq:conditional_Gaussian} is compatible and the joint distribution of $X_{\col}$ is of the form
  $
  p(x_{\col}) \propto \exp \big(-\frac{1}{2} x_{\col}^\top \cdot  \Sigma_{\col}^{-1} \cdot x_{\col}\big)
  $
  for some $ \Sigma_{\col} \in \RR^{pm \times pm}$.
  When the choice of $M$ and $\Sigma_j$ is such that $ \Sigma_{\col} \succ 0$, the conditional distributions are strongly compatible and we have $ X_{\col} \sim N(0, \Sigma_{\col})$. 
  
%  The conditional distribution \eqref{eq:conditional_Gaussian} defines
%  a valid joint distribution $ X_{\col} \sim N(0, \Sigma_{\col})$ for
%  some $ \Sigma_{\col} \in \RR^{pm \times pm}$ and for $M$ and
%  $ \Sigma_j$ such that $ \Sigma_{\col} \succ 0$.
\end{proposition}

\vspace{-3mm}
The explicit expression of $ \Sigma_{\col}$ can be derived from
\eqref{eq:conditional_Gaussian}, however, in general is quite
complicated. 
The following example provides an explicit formula in the case where $\Sigma_j = I$.

\begin{example}
\label{example:simple}
Suppose that $ \Sigma_j = I$ for all $j=1,\ldots,m$. Let
$A \in \RR^{m \times m}$ denote the adjacency matrix of the graph $G$,
with $a_{j,k} = 1$ when there is an edge between nodes $j$ and $k$ and
$0$ otherwise. Denote $\ell^c = \{1, \ldots,\ell-1,\ell+1,\ldots,p\}$, the conditional distribution of  $[ x_j]_\ell$ is given by
\vspace{1mm}
\[
[ x_j]_\ell \, \Big| \, x_{c_j}, [ x_j]_{\ell^c} \sim N\bigg( \Big[  M\sum_{k\in c_j}  x_k \Big]_\ell, 1 \bigg).
%= N\bigg(  M_{\ell,:} \cdot \sum_{k\in c_j}  x_k, 1 \bigg)
\]
Moreover, there exists a joint distribution
$ X_{\col} \sim N(0, \Sigma_{\col})$ where
$ \Sigma_{\col} \in \RR^{pm \times pm}$ satisfies
%is a block diagonal matrix with $m$ blocks satisfying 
%$[ \Sigma_{\col}^{-1}]_{i,i} = I$ and $[ \Sigma_{\col}^{-1}]_{i,j} = -M$ iff $a_{i,j} = 1$. Here the subscript $(i, j)$ is for the $i^{th}$ $\RR^{m \times m}$ block. 
%In matrix form, we have the Kronecker product
\#
 \Sigma_{\col}^{-1} = I - A \otimes M,
\#
and $A \otimes M$ denotes the Kronecker product between $A$ and $M$.
Clearly, we need $\Sigma_{\col} \succ 0$, which imposes implicit restrictions
on $A$ and $M$.
To ensure that $ \Sigma_{\col}$ is positive definite, we need to
ensure that all the eigenvalues of $A \otimes M$ are smaller than
$1$. One sufficient condition for this is  $\|A\|_2 \cdot \|M\|_2 < 1$.
For example, consider a chain graph with
\vspace{1mm}
\#
\label{eq:expression_chain}
A = 
\begin{bmatrix}
0 & 1 && \\
1 & 0 & \ddots & \\
& \ddots & \ddots & 1 \\
&& 1 & 0
\end{bmatrix} \in \RR^{p \times p}
\,\,\, \text{and} \,\,\,
 \Sigma_{\col}^{-1} = 
\begin{bmatrix}
I & -M & & \\
-M & I & \ddots & \\
& \ddots & \ddots & -M \\
& & -M & I
\end{bmatrix} \in \RR^{pm \times pm}.
\#
Then it suffices to have $\| M \|_2 < 1/2$. Similarly for $\omega$-nearest
neighbor structure, it suffices to have $\| M \|_2 < 1/2\omega$ and for
the lattice structure to have $\| M \|_2 < 1/4$.
\end{example}

\subsection{Estimation Procedures}

% The density of obtaining $ x_j$ conditioning on $ x_{c_j}$ is
% \begin{equation}
% \PP[ x_j |  x_{c_j},  M] = \Big[ \det(2\pi \Sigma_j) \Big]^{-\frac 12}\cdot \exp\bigg[-\frac{1}{2}\cdot \Big(  x_j-\sum_{k\in c_j} M x_k \Big)^\top  \Sigma_j^{-1} \Big(  x_j-\sum_{k\in c_j} M x_k \Big)  \bigg].
% \end{equation}
Since $ \Sigma_j$ is unknown, we propose to minimize the following 
loss function based on the conditional log-likelihood
\begin{equation}
\label{eq:log_likeli}
\cL( M) = m^{-1}\sum_{j=1}^m\cL^{j}( M),
\end{equation}
where
$\cL^j( M):= \frac{1}{2}\cdot\big\| x_j- M\sum_{k\in c_j}
x_k\big\|^2$.
Let $ M^* = { V^*} { V^*}^\top$ denote the true rank $r$ matrix with
$ V^* \in \RR^{p \times r}$. Note that $V^*$ is not unique, but $M^*$
is.  Observe that minimizing \eqref{eq:log_likeli} leads to a
consistent estimator, since
%the gradient 
%\[
%\nabla \cL( M) = -\frac{1}{2m}\cdot\sum_{j=1}^m\bigg[\Big( x_j -  M\sum_{k\in c_j} x_k\Big)\cdot\sum_{k\in c_j} x_k^\top + \sum_{k\in c_j} x_k\cdot\Big( x_j -  M\sum_{k\in c_j} x_k\Big)^\top\bigg], 
%\]
%and thus
\begin{align*}
\EE\Big[\nabla \cL^j( M^*)\Big] = 
\EE\bigg[\Big( x_j -  M^*\sum_{k\in c_j} x_k\Big)\sum_{k\in c_j} x_k^\top\bigg] &=
\EE_{ x_{c_j}}\EE_{ x_j}\bigg[\Big( x_j -  M^*\sum_{k\in c_j} x_k\Big)\sum_{k\in c_j} x_k^\top~\Big|  x_{c_j}\bigg] = 0.
\end{align*}
In order to find a low rank solution $\hat M$ that approximates $M^*$, we
consider the following two procedures.

\paragraph{Convex Relaxation}

We solve the following problem
\begin{align}
\label{eq:optimization_convex}
\min_{ M\in \RR^{p\times p},  M^\top =  M,  M  \succeq 0} \cL( M) +\lambda\| M\|_{*},
\end{align}
where $\| \cdot \|_*$ is the nuclear norm of a matrix and $\lambda$ is
the regularization parameter to be specified in the next section. The
problem \eqref{eq:optimization_convex} is convex and hence can be
solved by proximal gradient descent method \cite{parikh2014proximal}
with any initialization point.

\paragraph{Non-convex Optimization}

Although it is guaranteed to find global minimum by solving the convex
relaxation problem \eqref{eq:optimization_convex}, in practice it may
be slow.  In our problem, since $ M$ is low rank and positive
semidefinite, we can always write $ M = V V^\top$ for some
$ V \in \RR^{p \times r}$ and solve the following non-convex problem
\begin{align}
\label{eq:optimization_nonconvex}
\min_{ V\in \RR^{p\times r}} \cL( V V^\top).
\end{align}
With an appropriate initialization $ V^{(0)}$, in each iteration we
update $ V$ by gradient descent
\[
 V^{(t+1)} =  V^{(t)} - \eta \cdot \left.\nabla_{ V} \cL( V V^\top)  \right\vert_{ V =  V^{(t)}},
\]
where $\eta$ is the step size. The choice of initialization $ V^{(0)}$
and step size $\eta$ will be specified in details in the next
section. The unknown rank $r$ can be estimated as in
\cite{bunea2011optimal}.

\section{Theoretical Result}
\label{sec:theoretical}

We establish convergence rates for the two estimation procedures.

\subsection{Convex Relaxation}
\label{subsec:convex}

In order to show that a minimizer of \eqref{eq:optimization_convex}
gives a good estimator for $M$, we first show that the objective
function $\cL(\cdot)$ is strongly convex under the assumption that the 
data are distributed according to \eqref{eq:conditional_Gaussian} with the true parameter
 $ M^* = { V^*} { V^*}^\top$ with $V^* \in \RR^{p \times r}$. Let 
\begin{align*}
\delta\cL( \Delta) &= \cL( M^* + \Delta) - \cL( M^*) - \langle\nabla\cL( M^*),  \Delta\rangle,
\end{align*}
where $\langle A, B\rangle=\tr(A^\top B)$ and $\Delta$ is a symmetric
matrix.  Let $\Delta_i$ denote the $i$-th column of $\Delta$ and let
$\Delta_{\col} = [\Delta_1^\top, \ldots, \Delta_p^\top]^\top \in
\RR^{p^2 \times 1}$
be the vector obtained by stacking columns of $\Delta$. Then a simple
calculation shows that
\[
\begin{aligned}
\delta\cL( \Delta) &= \frac{1}{m}\cdot \sum_{j=1}^m \big\| \Delta\sum_{k\in c_j} x_k\big\|^2 
%&= \frac{1}{m}\cdot \sum_{j=1}^m \bigg\| 
%\begin{pmatrix}
% \Delta_1^\top\sum_{k\in c_j} x_k \\
%\vdots \\
% \Delta_p^\top\sum_{k\in c_j} x_k 
%\end{pmatrix}
%\bigg\|^2 \\
%= \frac{1}{m}\cdot \sum_{j=1}^m \sum_{i=1}^p \big(  \Delta_i^\top\sum_{k\in c_j} x_k \big)^2 \\
%&= \frac{1}{m}\cdot \sum_{j=1}^m \sum_{i=1}^p  \Delta_i^\top \big(\sum_{k\in c_j} x_k\big) \cdot \big(\sum_{k\in c_j} x_k\big)^\top   \Delta_i
= \sum_{i=1}^p  \Delta_i^\top \bigg[ \frac{1}{m} \sum_{j=1}^m \big(\sum_{k\in c_j} x_k\big) \cdot \big(\sum_{k\in c_j} x_k\big)^\top \bigg]     \Delta_i
\end{aligned}
\]
has a quadratic form in each $\Delta_i$ with the same Hessian matrix $H$. Let
\[
\begin{aligned}
\tilde  X &= \Big[ \sum_{k\in c_1} x_k, \sum_{k\in c_2} x_k, ..., \sum_{k\in c_m} x_k \Big] =  X \cdot A \in \RR^{p \times m},
\end{aligned}
\]
where $A$ is the adjacency matrix of the graph $G$. Then the Hessian matrix is given by
\#
\label{eq:def_H}
 H = \frac{1}{m} \sum_{j=1}^m \big(\sum_{k\in c_j} x_k\big) \cdot \big(\sum_{k\in c_j} x_k\big)^\top = \frac{1}{m} \tilde  X \tilde  X^\top = \frac 1m  X A A^\top  X^\top \in \RR^{p \times p}
\#
and therefore we can succinctly  write $
\delta\cL( \Delta) =  \Delta_{\col}^\top \cdot  H_{\col} \cdot  \Delta_{\col}$, 
where the total Hessian matrix $ H_{\col} = \diag( H, H, \ldots, H ) \in \RR^{p^2 \times p^2}$ is a block diagonal matrix.

To show that $\cL(\cdot)$ is strongly convex, it suffices to lower
bound the minimum eigenvalue of $ H$, defined in \eqref{eq:def_H}.  If
the columns of $\tilde X$ were independent, the minimum eigenvalue of
$ H$ would be bounded away from zero with overwhelming probability for
a large enough $m$ \cite{raskutti2010restricted}. However, in our
setting the columns of $\tilde X$ are dependent and we need to prove
this lower bound using different tools.  As the distribution of $ X$
depends on the unknown conditional covariance matrices $ \Sigma_j$,
$j=1,\ldots,m$ in a complicated way, we impose the following
assumption on the expected version of $ H$.

\paragraph{Assumption EC.}

The minimum and maximum eigenvalues of
$\EE H$ are bounded from below and from above:
$0 < c_{\min} \leq \sigma_{\min} (\EE H) \leq \sigma_{\max} (\EE H) \leq
c_{\max} < \infty$.

Assumption (EC) puts restrictions on conditional covariance matrices  $\Sigma_j$
and can be verified in specific instances of the problem. In the context
of Example~\ref{example:simple}, where $ \Sigma_j = I$, $j=1,\ldots,m$,
and the graph is a chain, we have the adjacency matrix $A$
and the covariance matrix $ \Sigma_{\col}$ given in~\eqref{eq:expression_chain}.
Then simple linear algebra \cite{hu1996analytical} gives us that
\[
\EE  H = m^{-1} \EE X AA^\top X^\top = 2 I + c M^2 + o( M^2),
\]
which guarantees that $ \sigma_{\min} (\EE H)\geq 1$ and
$\sigma_{\max} (\EE H)\leq c+3$ for large enough $m$.

The following assumption requires that the spectral norm of $A$ and
$ \Sigma_{\col}$ do not scale with $m$.

\paragraph{Assumption SC.} 

There exists a constant $\rho_0$ such that $\max\big\{ \|A\|_2, \| \Sigma_{\col}^{1/2}\|_2 \big\} \leq \rho_0$. 

Assumption (SC) gives sufficient condition on the graph structure, and
it requires that the dependency among nodes is weak. In fact, it can be
relaxed to $\rho_0 = o( m^{1/4} )$ which allows the spectral norm to
scale with $m$ slowly. In this way, the minimum and maximum eigenvalues
in assumption (EC) also scale with $m$ and it results in a much larger
sample complexity on $m$. However, if $\rho_0$ grows even faster, then
there is no way to guarantee a reasonable estimation.  We see that
$\rho_0 \sim m^{1/4}$ is the critical condition, and we have the phase
transition on this boundary.

%\marginpar{add a paragraph to clarify that the assumptions are not restrictive}
It is useful to point out that these assumptions are not restrictive. For example, under the simplification that $\Sigma_j = I$, we have $\| \Sigma_{\col} \|_2 = 1/(1 - \|A\|_2 \cdot \| M \|_2 )$. The condition $\|A\|_2 \cdot \| M \|_2 < 1$ is satisfied naturally for a valid Gaussian embedding model. Therefore in order to have $\| \Sigma_{\col}^{1/2}\|_2 \leq \rho_0$, we only need that $\|A\|_2 \cdot \| M \|_2  \leq 1-1/\rho_0^2$, i.e., it is bounded away from 1 by a constant distance.

%Assumption (SC) requires that the dependency among nodes is weak.  
It is straightforward to verify that assumption (SC) holds for the chain
structure in Example~\ref{example:simple}. If the graph is
fully connected, we have $\|A\|_2 = m-1$, which violates the
assumption. In general, assumption (SC) gives a sufficient condition
on the  graph structure so that the embedding is learnable.

With these assumptions, the following lemma proves that the minimum
and maximum eigenvalues of the sample Hessian matrix $ H$ are also
bounded from below and above with high probability.

\begin{lemma}
\label{lemma:sigma_min(H)}
Suppose the assumption (EC) and (SC) hold. Then 
for $m \geq c_0 p$ we have $ \sigma_{\min} ( H) \geq \frac{1}{2} c_{\min}$ 
and $ \sigma_{\max} ( H) \leq 2 c_{\max}$ with probability at least
$1-c_1\exp(-c_2m)$, where $c_0, c_1, c_2$ are absolute constants. Therefore 
\#
\kappa_{\mu}\cdot \| \Delta\|_{F}^2 \leq \delta\cL( \Delta) \leq \kappa_{L}\cdot \|\Delta\|_{F}^2,
\# 
with $\kappa_{\mu} = \frac{1}{2} c_{\min}$ 
and $\kappa_L = 2 c_{\max}$ 
for any $ \Delta \in \RR^{p \times p}$.
\end{lemma}

Lemma \ref{lemma:sigma_min(H)} is the key technical result, which
shows that although all the $x_j$ are dependent, the objective
function $\cL(\cdot)$ is still strongly convex and smooth in $\Delta$.
Since the loss function $\cL(\cdot)$ is strongly convex, an 
application of Theorem~1 in \cite{negahban2011estimation} gives the
following result on the performance of the convex relaxation approach proposed in the
previous section.

\begin{theorem}
\label{theorem:convex}
Suppose the assumptions (SC) and (EC) are satisfied. The minimizer  $\hat{ M}$
of \eqref{eq:optimization_convex}  with 
$
\lambda \ge \Big\|\frac{1}{m} \sum_{j=1}^m\Big( x_j -  M^*\sum_{k\in c_j} x_k\Big)\cdot\sum_{k\in c_j} x_k^\top\Big\|_2
$
satisfies
\[
    \|\hat{ M} -  M^*\|_{F}\le \frac{32\sqrt{r}\lambda}{\kappa_\mu}.
\]
\end{theorem}
\vspace{-2mm}
The following lemma gives us a way to set the regularization parameter $\lambda$.
\begin{lemma}
\label{lemma:lambda}
Let $ G = \frac{1}{m} \sum_{j=1}^m \Sigma_j$. Assume that the maximum
eigenvalue of $ G$ is bounded from above as
$\sigma_{\max}( G) \leq \eta_{\max}$ for some constant
$\eta_{\max}$. Then there exist constants $c_0, c_1, c_2, c_3 > 0$ such
that for $m \geq c_0 p$, we have
\[
\PP \Bigg[ \,
 \bigg\|\frac{1}{m} \sum_{j=1}^m\Big( x_j -  M^*\sum_{k\in c_j} x_k\Big)\cdot\sum_{k\in c_j} x_k^\top\bigg\|_2
  \geq c_1 \sqrt{\frac{p}{m}} \, \Bigg] \leq c_2 \exp(-c_3m).
\]
\end{lemma}
\vspace{-2mm}
Combining the result of Lemma \ref{lemma:lambda} with 
Theorem \ref{theorem:convex}, we see that $\lambda$ should be chosen as
$\lambda = \cO \big( \sqrt{p/m} \big)$, 
which leads to the error rate %\mcomment{Why is this optimal?}
\begin{equation}
\label{eq:convex_rate}
\|\hat{ M} -  M^*\|_{F} = \cO_P \rbr{ \frac{1}{\kappa_\mu} \sqrt{\frac{rp}{ m}} }.
\end{equation}

\subsection{Non-convex Optimization}
\label{subsec:nonconvex}

Next, we consider the convergence rate for the non-convex method
resulting in minimizing \eqref{eq:optimization_nonconvex} in
$V$. Since the factorization of $ M^*$ is not unique, we measure the
subspace distance between $ V$ and $ V^*$.

\vspace{-3mm}
\paragraph{Subspace distance.} 
Let $ V^*$ be such that $ V^*{V^*}^\top = \Theta^*$. 
Define the subspace distance between
$ V$ and
$ V^*$ as
\begin{equation}
d^2(V,V^*) = \min_{O \in \cO(r)} \|V - V^*O \|_F^2,
\end{equation}
where $\cO(r) = \{O: O \in \RR^{r \times r}, OO^\top = O^\top O = I \}$.

Next, we introduce the notion of the statistical error. Denote %\mcomment{Is $\Delta$ symmetric?}
\begin{equation*}
\Omega = \big\{  \Delta:  \Delta \in \RR^{p \times p}, \Delta = \Delta^\top, \rank( \Delta) = 2r, \| \Delta\|_F = 1 \big\}.
\end{equation*}
The statistical error is defined as
\begin{equation}
\begin{aligned}
e_{\text{stat}} &= \sup_{ \Delta \in \Omega} \, \big\langle \nabla\cL( M^*),  \Delta \big\rangle.
\end{aligned}
\end{equation}
Intuitively, the statistical error quantifies how close the estimator
can be to the true value. Specifically, if $V$ is within
$c \cdot e_{\text{stat}}$ distance from $V^*$, then it is 
already optimal. For any $ \Delta \in \Omega$, we have the
factorization $ \Delta = U_{ \Delta}V_{ \Delta}^\top$ where
$U_{ \Delta}, V_{ \Delta} \in \RR^{p \times 2r}$ and
$\|U_{ \Delta}\|_2 = \|V_{ \Delta}\|_F = 1$. We then have
\begin{equation}
\begin{aligned}
\label{eq:stat_error_rate}
\big\langle \nabla\cL( M^*),  \Delta \big\rangle 
%&= \big\langle \nabla\cL( M^*), U_{ \Delta}V_{ \Delta}^\top \big\rangle 
&= \big\langle \nabla\cL( M^*) V_{ \Delta}, U_{ \Delta} \big\rangle \leq \| \nabla\cL( M^*) V_{ \Delta}\|_F \cdot \|U_{ \Delta} \|_F \\
&\leq \| \nabla\cL( M^*)\|_2 \|V_{ \Delta}\|_F  \|U_{ \Delta} \big\|_F \leq \sqrt{2r}\lambda, 
\end{aligned}
\end{equation}
where the last inequality follows from Lemma~\ref{lemma:lambda}. In
particular, we see that both convex relaxation and non-convex
optimization give the same rate.

\vspace{-1mm}
\paragraph{Initialization.} In order to prove a linear rate of
convergence for the procedure, we need to initialize it properly.
Since the loss function $\cL( M)$ is quadratic in $ M$, we can 
ignore all the constraints on $ M$ and get a closed form solution as
\#
 M^{(0)} = \Big[ \sum_{j=1}^m \big(\sum_{k\in c_j} x_k\big)\big(\sum_{k\in c_j} x_k\big)^\top \Big]^{-1} \cdot \Big[ \sum_{j=1}^m  x_j^\top \big(\sum_{k\in c_j} x_k\big) \Big].
\#
We then apply rank-$r$ eigenvalue decomposition on 
$\tilde M^{(0)} = \frac{1}{2}\big(M^{(0)} + {M^{(0)}}^\top\big)$ 
and obtain 
$[\tilde V, \tilde S, \tilde V] = {\text{rank-}}r { \text{ svd of }} \tilde M^{(0)}$. 
Then $V^{(0)} = \tilde V \tilde S^{1/2}$ is the initial point for the 
gradient descent. The following lemma quantifies the accuracy of this initialization.

\begin{lemma}
\label{lemma:initialization}
The initialization $ M^{(0)}$ and $V^{(0)}$ satisfy %\mcomment{Maybe say also how close initial $V$ is?}
\[
\| M^{(0)} -  M^{*}\|_F \leq \frac{2\sqrt{p}\lambda}{\kappa_\mu} \,\,\,\, \text{and} \,\,\,\, d^2 \big( V^{(0)}, V^* \big) \leq\frac{20p\lambda^2}{\kappa_\mu^2 \cdot \sigma_r(M^*)}
\]
where $ \sigma_r(M^*)$ is the minimum non-zero singular value of $M^*$.
\end{lemma}

With this initialization, we obtain the following main result for the
non-convex optimization approach, which establishes a linear rate of
convergence to a point that has the same statistical error rate as the
convex relaxation approach studied in Theorem \ref{theorem:convex}.

\begin{theorem}
\label{theorem:nonconvex}
Suppose the assumption (EC) and (SC) are satisfied, and suppose the
step size $\eta$ satisfies
%\begin{equation}
%\label{eq:eta_constant}
$\eta \leq \big[ {32\|M^{(0)}\|_2^2} \cdot (\kappa_\mu + \kappa_L ) \big]^{-1}$. 
%$\eta \leq \frac{1}{{32\|M^{(0)}\|_2^2} \cdot (\kappa_\mu + \kappa_L )}$, 
%\end{equation}
For large enough $m$, after $T$ iterations we have
\begin{equation}
d^2 \Big( V^{(T)}, V^* \Big) \leq \beta^T d^2 \Big( V^{(0)}, V^* \Big) + \frac{C}{\kappa_\mu^2}\cdot e^2_{\rm{stat}}, 
\end{equation}
for some constant $\beta < 1$ and a constant $C$.
\end{theorem}

\section{Experiment}
\label{sec:experiment}

\begin{figure}
    \centering
    \subfigure[Chain structure]
    {
        \includegraphics[width=0.31\textwidth]{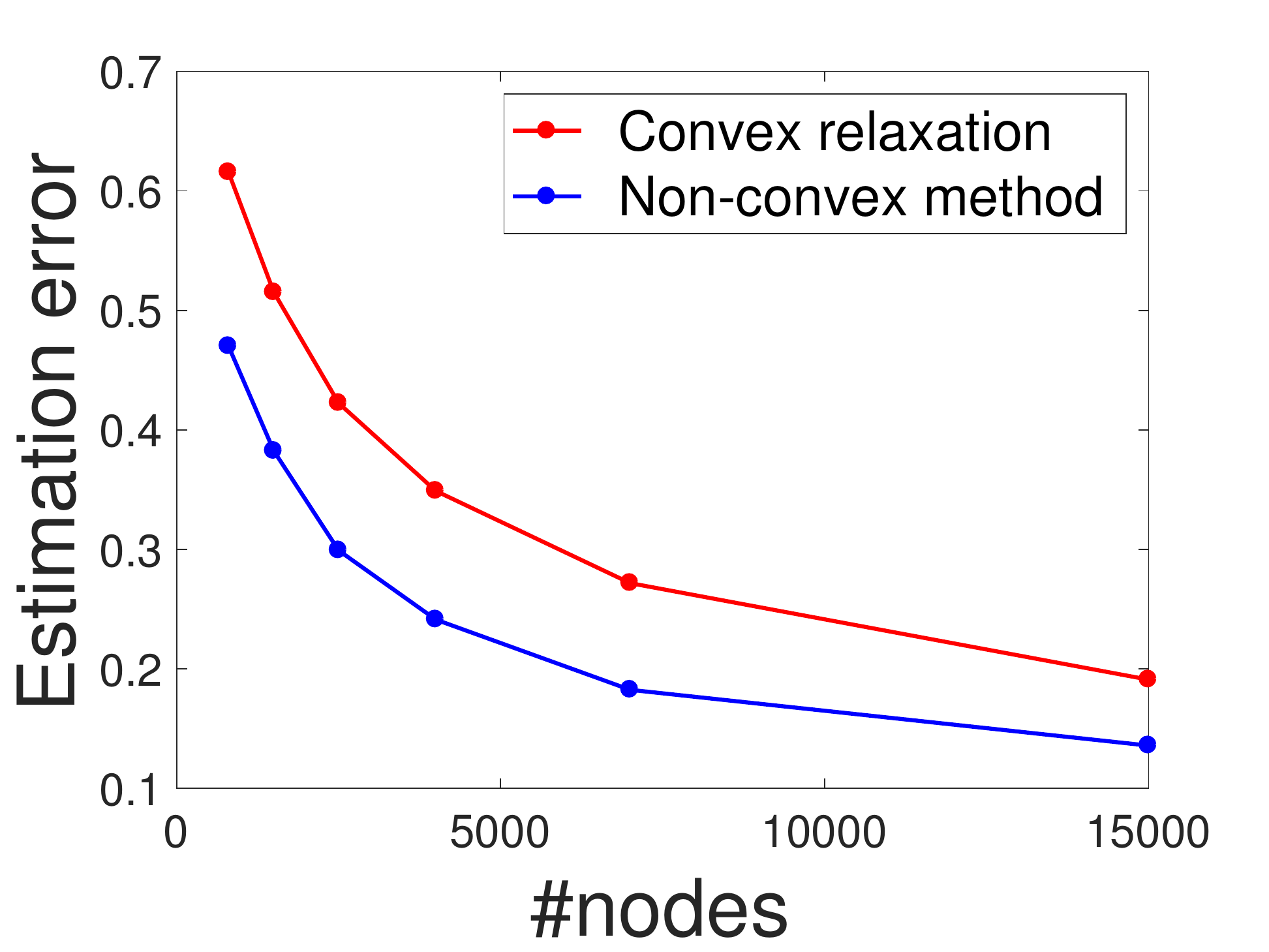}
        \label{fig:err_chain}
    }
    \subfigure[$\omega$-nearest neighbor structure]
    {
        \includegraphics[width=0.31\textwidth]{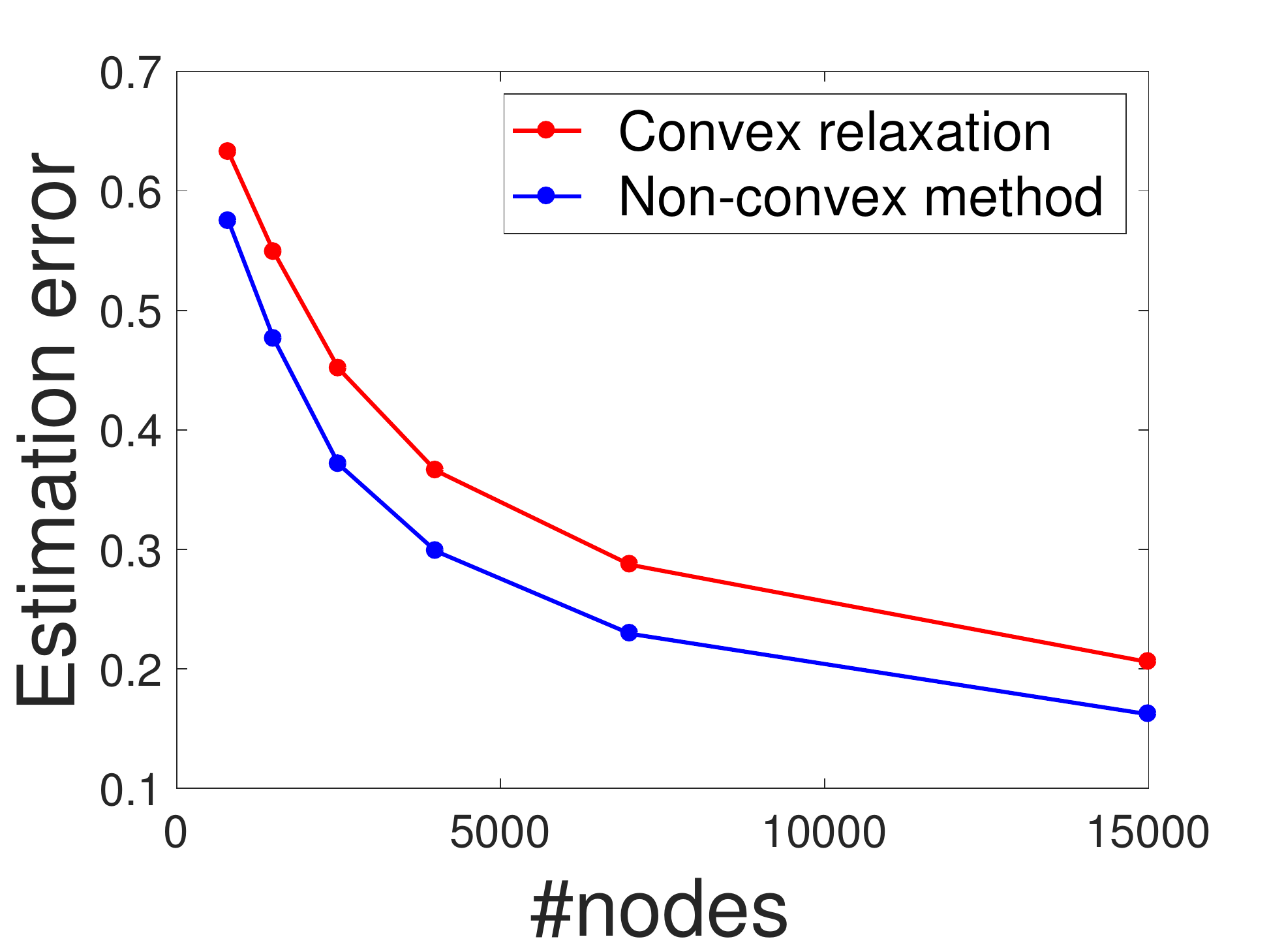}
        \label{fig:err_near}
    }
    \subfigure[Lattice structure]
    {
        \includegraphics[width=0.31\textwidth]{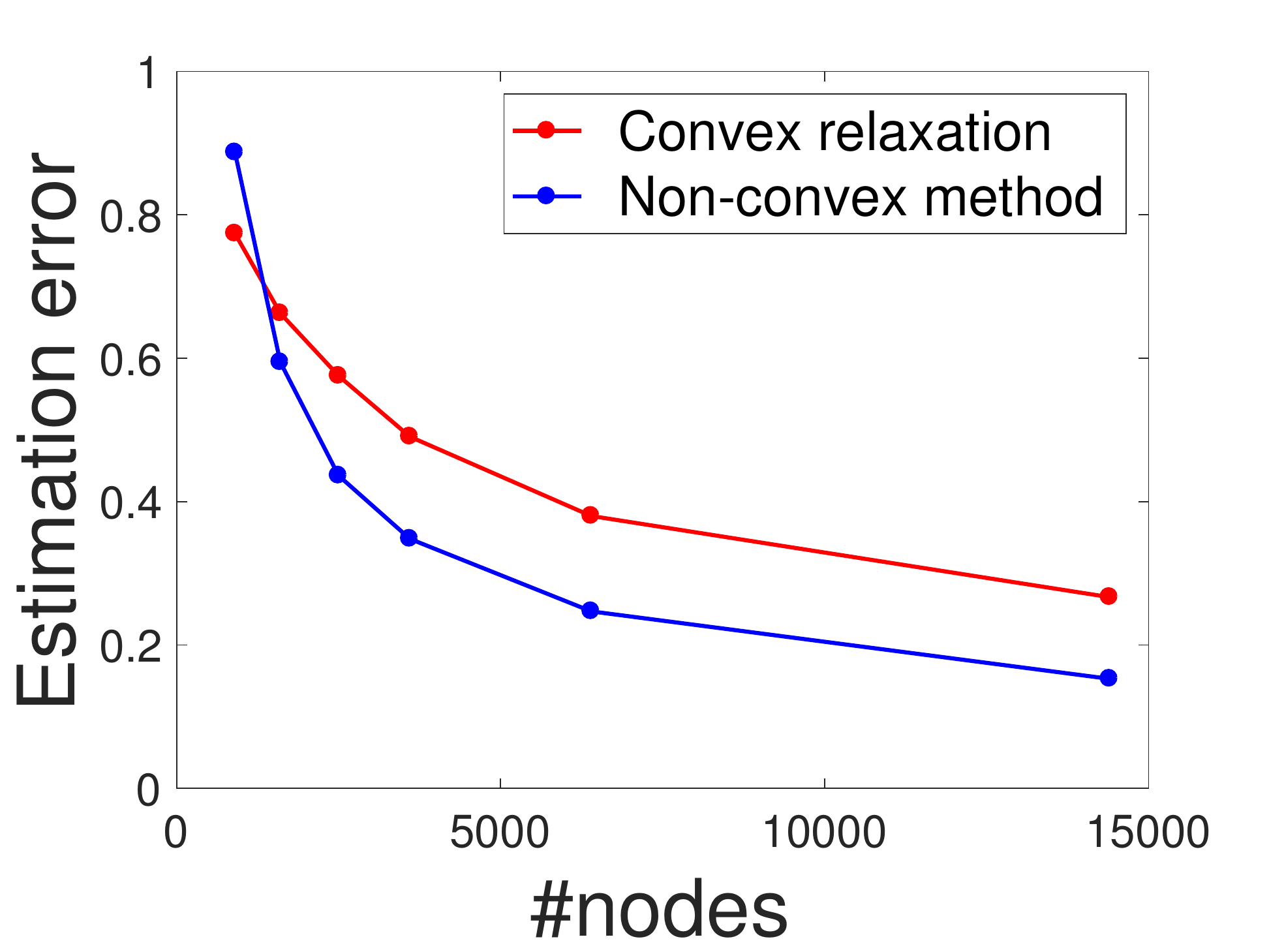}
        \label{fig:err_lattice}
    }
    \caption{Estimation accuracy for three context structures}
    \label{fig:err_context_structures}
\end{figure}

%\begin{figure*}[htbp]
%\begin{minipage}[t]{0.32\linewidth}
%\centering
%\includegraphics[width=0.95\textwidth]{err_chain}
%\caption{Estimation accuracy for chain structure}
%\label{fig:err_chain}
%\end{minipage}
%\begin{minipage}[t]{0.32\linewidth}
%\centering
%\includegraphics[width=0.95\textwidth]{err_near}
%\caption{Estimation accuracy for $\kappa$-nearest neighbor structure}
%\label{fig:err_near}
%\end{minipage}
%\begin{minipage}[t]{0.32\linewidth}
%\centering
%\includegraphics[width=0.95\textwidth]{err_lattice}
%\caption{Estimation accuracy for lattice structure}
%\label{fig:err_lattice}
%\end{minipage}
%\end{figure*}

\begin{figure}
    \centering
    \subfigure[Chain structure]
    {
        \includegraphics[width=0.31\textwidth]{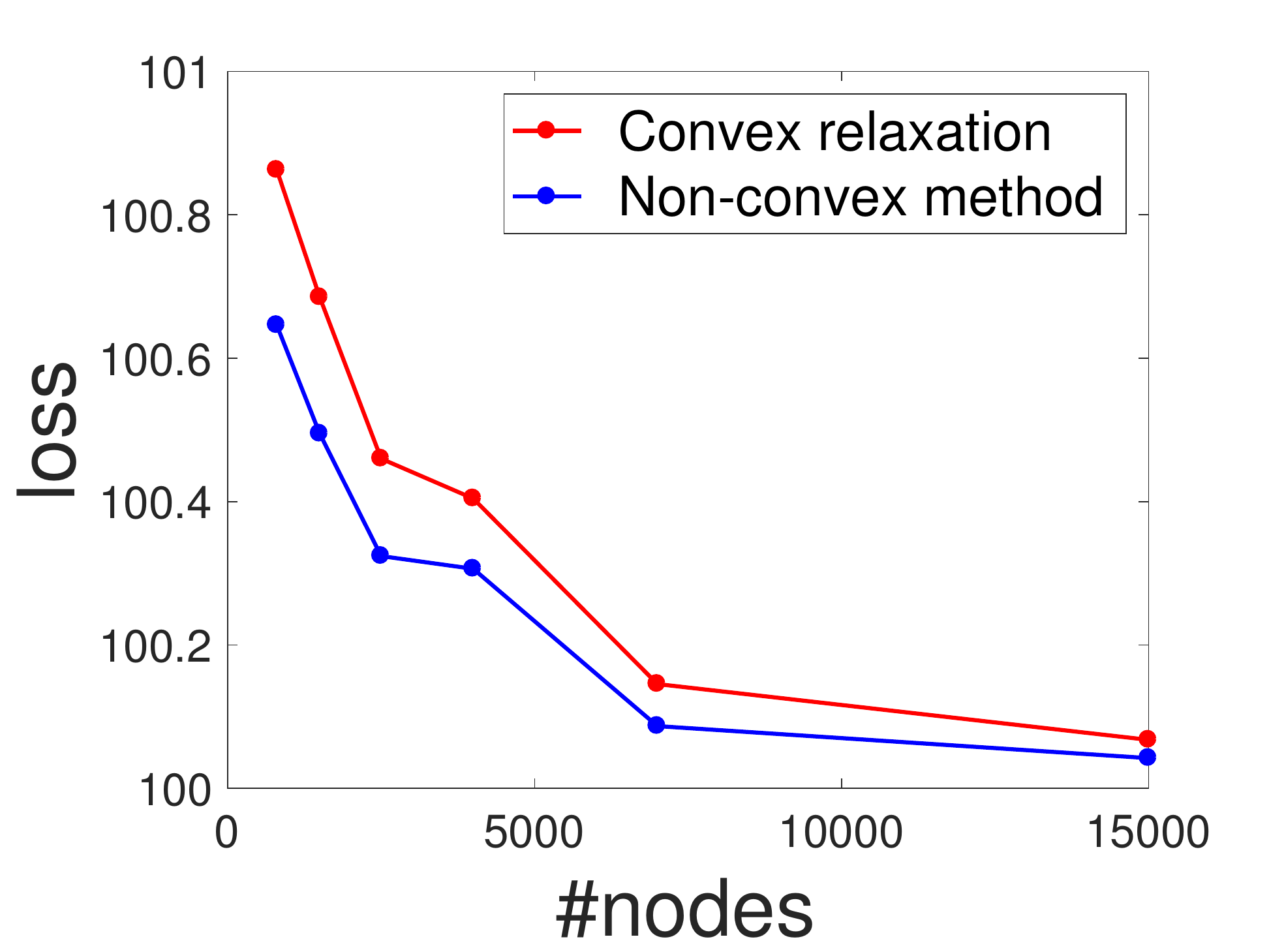}
        \label{fig:loss_chain}
    }
    \subfigure[$\omega$-nearest neighbor structure]
    {
        \includegraphics[width=0.31\textwidth]{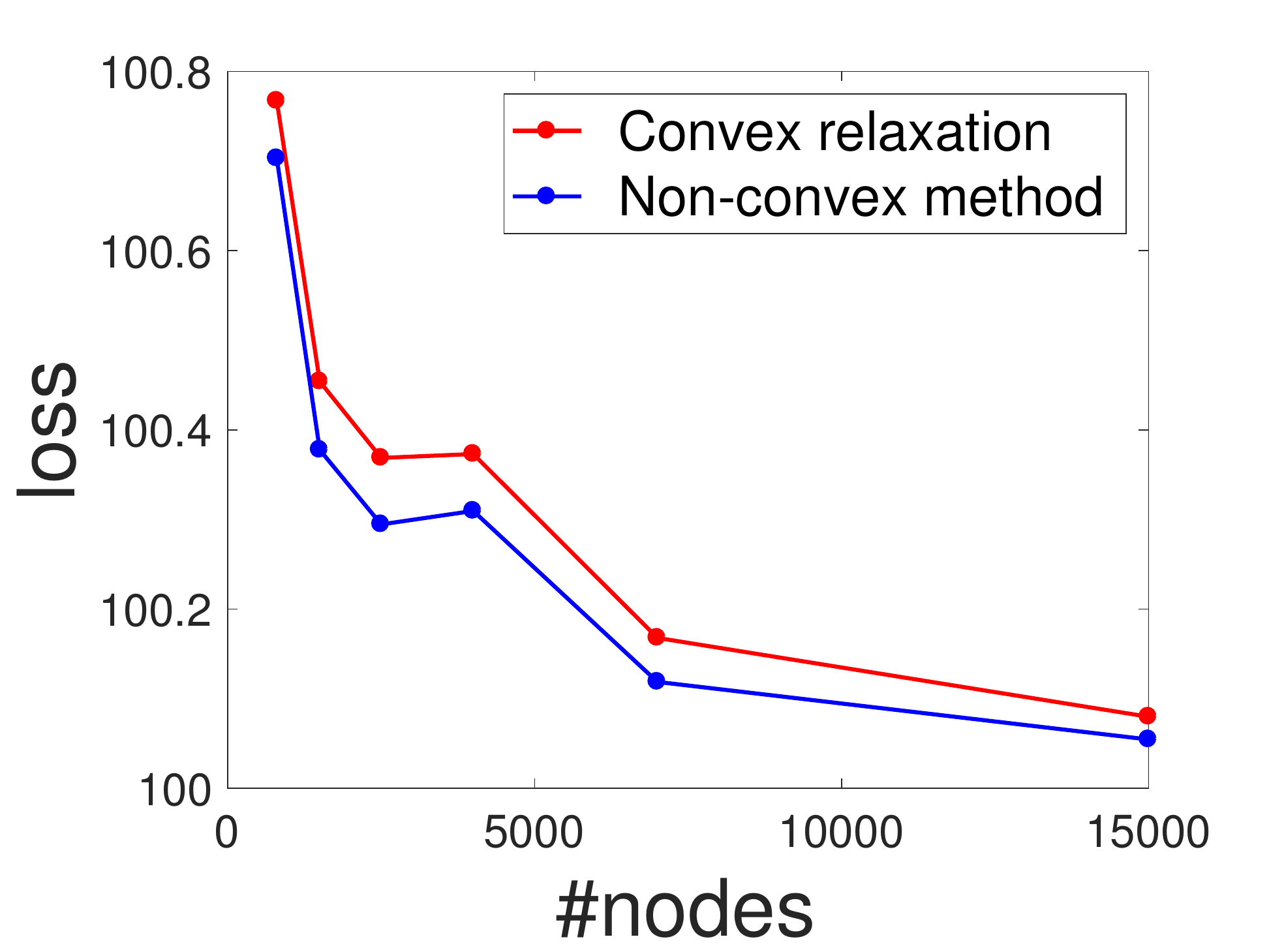}
        \label{fig:loss_near}
    }
    \subfigure[Lattice structure]
    {
        \includegraphics[width=0.31\textwidth]{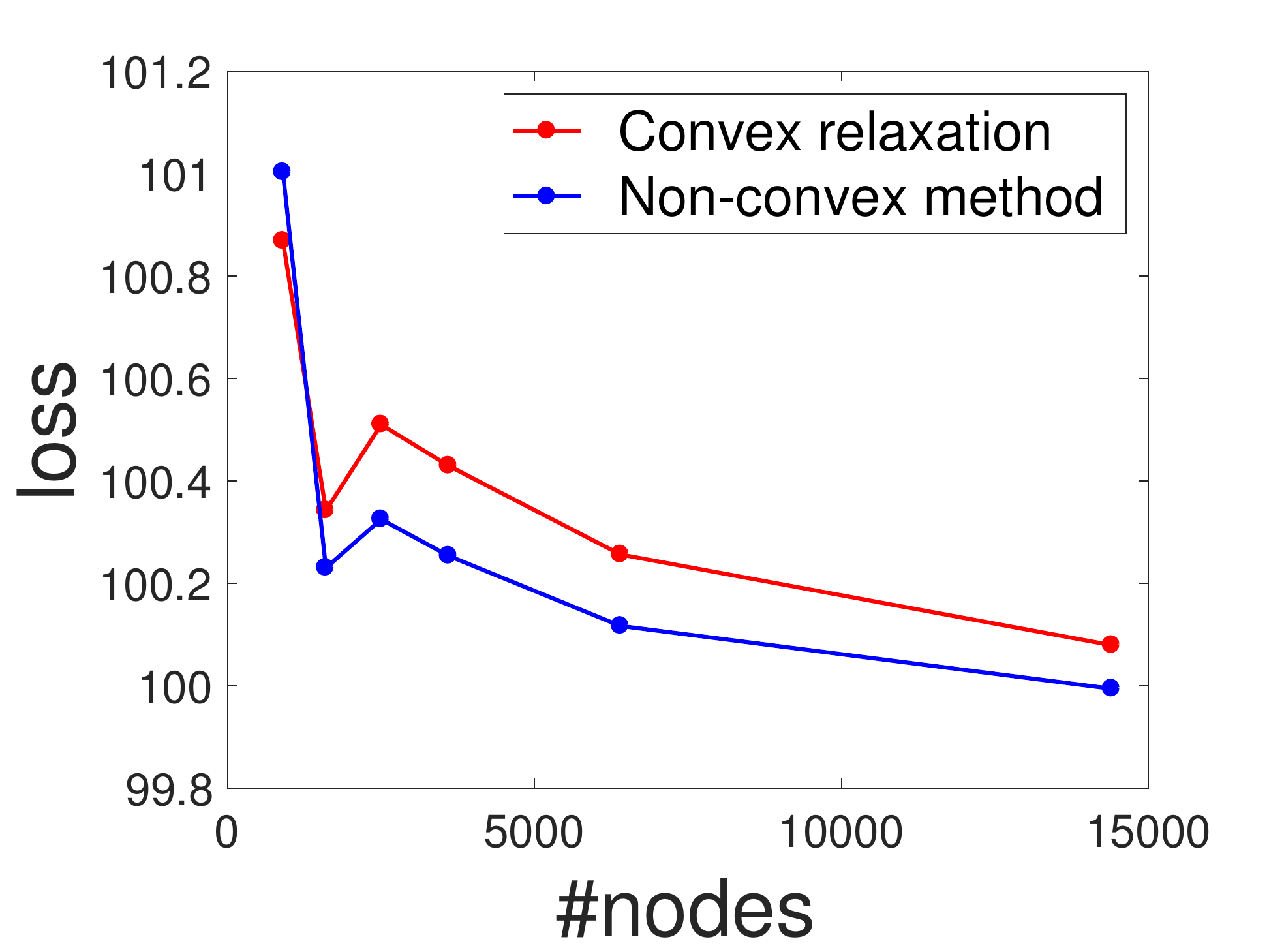}
        \label{fig:loss_lattice}
    }
    \caption{Testing loss for three context structures}
    \label{fig:loss_context_structures}
\end{figure}

%\begin{figure*}[htbp]
%\begin{minipage}[t]{0.32\linewidth}
%\centering
%\includegraphics[width=0.95\textwidth]{loss_chain}
%\caption{Testing loss for \newline chain structure}
%\label{fig:loss_chain}
%\end{minipage}
%\begin{minipage}[t]{0.32\linewidth}
%\centering
%\includegraphics[width=0.95\textwidth]{loss_near}
%\caption{Testing loss for \newline $\kappa$-nearest neighbor structure}
%\label{fig:loss_near}
%\end{minipage}
%\begin{minipage}[t]{0.32\linewidth}
%\centering
%\includegraphics[width=0.95\textwidth]{loss_lattice}
%\caption{Testing loss for \newline lattice structure}
%\label{fig:loss_lattice}
%\end{minipage}
%\end{figure*}

%\ref{fig:chain} - \ref{fig:lattice}
In this section, we evaluate our methods through experiments. 
%If we know the covariance matrix or have a good estimate of it, we could include that in the loss function (3.4). This may improve the estimation accuracy. However, even if they are unknown, minimizing the loss function (3.4) directly still gives a consistent estimation, as shown by the formula after line 153. 
%\marginpar{add experiments to justify unknown cov matrix}
We first justify that although $\Sigma_j$ is unknown, minimizing \eqref{eq:log_likeli} still leads to a consistent estimator. 
We compare the estimation accuracy with known and unknown covariance matrix $\Sigma_j$. We set $\Sigma_j = \sigma_j \cdot {\text{Toeplitz}}(\rho_j)$ where ${\text{Toeplitz}}(\rho_j)$ denotes Toeplitz matrix with parameter $\rho_j$. 
We set $\rho_j \sim U[0,0.3]$ and $\sigma_j \sim U[0.4, 1.6]$ to make them non-isotropic. 
The estimation accuracy with known and unknown $\Sigma_j$ are given in Table \ref{justify_cov}. 
We can see that although knowing $\Sigma_j$ could give slightly better accuracy, the difference is tiny. Therefore, even if the covariance matrices are not isotropic, ignoring them still gives a consistent estimator.

\begin{table}[htp]
\caption{Comparison of estimation accuracy with known and unknown covariance matrix}
\vspace{-1mm}
\begin{small}
\begin{center}
\begin{tabular}{c|ccccc}
& $p=1000$ & $p = 2500$ & $p = 5000$ & $p = 8000$ & $p = 15000$  \\\hline
unknown &     0.8184  &  0.4432   & 0.3210    &0.2472  &  0.1723 \\
known &     0.7142  &  0.3990  &  0.2908  &  0.2288   & 0.1649  \\
\end{tabular}
\end{center}
\label{justify_cov}
\end{small}
\vspace{-2mm}
\end{table}%

We then consider three kinds of graph structures given in Figure \ref{fig:context_structures}: chain structure, $\omega$-nearest neighbor structure, and lattice structure. 
We generate the data according to the conditional distribution \eqref{eq:conditional_Gaussian} using Gibbs Sampling. We set $p=100, r=5$ and vary the number of nodes $m$. For each $j$, we set $ \Sigma_j =  \Sigma$ to be a Toeplitz matrix with $\Sigma_{i\ell} = \rho^{|i-\ell|}$ with $\rho = 0.3$. 
We generate independent train, validation, and test sets. 
For convex relaxation, the regularization parameter is selected using the validation set.
We consider two metrics, one is the estimation accuracy $\|\hat  M -  M^* \|_F / \| M^* \|_F$, and the other is the loss $\cL(\hat  M)$ on the test set. 

The simulation results for estimation accuracy for the three graph structures are shown in Figure \ref{fig:err_context_structures}, 
%\ref{fig:chain} - \ref{fig:lattice}, 
and the results for loss on test sets are shown in Figure \ref{fig:loss_context_structures}. 
%\ref{fig:err_chain} - \ref{fig:err_lattice}. 
Each result is based on 20 replicates. 
For the estimation accuracy, we see that when the number of nodes is small, neither method gives accurate estimation; for reasonably large $m$, non-convex method gives better estimation accuracy since it does not introduce bias; for large enough $m$, both methods give accurate and similar estimation. 
For the loss on test sets, we see that in general, both methods give smaller loss as $m$ increases. The non-convex method gives marginally better loss. This demonstrates the effectiveness of our methods.

\section{Conclusion}
\label{sec:conclusion}

%\marginpar{mention that the framework is for general model, just hard in probability perspective}
In this paper, we focus on Gaussian embedding and develop the first theoretical result for exponential family embedding model. We show that for various kinds of context structures, we are able to learn the embedding structure with only one observation. 
Although all the data we observe are dependent, we show that the objective function is still well-behaved and therefore we can learn the embedding structure reasonably well. 

It is useful to point out that, the theoretical framework we proposed is for {{\emph{general exponential family embedding models}}. As long as the similar conditions are satisfied, the framework and theoretical results hold for any general exponential family embedding model as well. 
However, proving these conditions is quite challenging from the probability perspective. Nevertheless, our framework still holds and all we need are more complicated probability tools. 
Extending the result to other embedding models, for example the Ising model, is work in progress.

\newpage
\bibliographystyle{plain}
\bibliography{GaussianEmbedding.bib}

\newpage
\appendix

\section{Technical proofs}

\subsection{Proof of Lemma \ref{lemma:sigma_min(H)}.}

\begin{proof}
Before we proceed with the main proof, we first introduce the following lemma in \cite{ledoux2013probability}. 
\begin{lemma}
\label{lemma:Talagrand}
Let $x_1, ..., x_T$ be independent and identically drawn from distribution $N(0,1)$ and $X = (x_1, ..., x_T)^\top$ be a random vector. Suppose a function $f: \RR^T \to \RR$ is Lipschitz, i.e., for any $v_1, v_2 \in \RR^T$, there exists $L$ such that $| f(v_1) - f(v_2) | \leq L \|v_1 - v_2 \|_2$, then we have that
\[
\PP\Big\{ |f(X) - \EE f(X) | > t \Big\} \leq 2\exp\Big(-\frac{t^2}{2L^2}\Big)
\]
for all $t > 0$.
\end{lemma}

We then proceed with the proof of Lemma \ref{lemma:sigma_min(H)}. 
For any fixed $ v \in \RR^p$ with $\| v\|_2 = 1$, define 
\[
W = f_v(Z) = \frac{1}{\sqrt m} \Big\|  v^\top \mat \big(  \Sigma_{\col}^{1/2} Z \big) \cdot A \Big\|_2,
\]
where $ Z \in \RR^{pm \times 1}$ and $\mat(\cdot)$ is a reshape operator that reshape a $pm$-dimensional vector to a $p \times m$ dimensional matrix. 
When $ Z \sim N(0,  I_{pm})$, 
it is straightforward to see that the distribution of $\mat \big(  \Sigma_{\col}^{1/2} Z \big)$ is the same as $ X$ and hence $W^2$ has the same distribution with $ v^\top  H  v$. 
We then verify that the function $f_v$ is Lipschitz with $L = \frac{\rho_0^2}{\sqrt m}$ where $\rho_0$ is defined in assumption (SC).
% = \|A\|_2\cdot \| \Sigma_{\col}^{1/2} \|_2$. 
For any vector $ Z_1,  Z_2$, we have
\begin{equation}
\begin{aligned}
\Big| f_v( Z_1) - f_v( Z_2) \Big| &= \frac{1}{\sqrt m} \bigg| \Big\|  v^\top re \big(  \Sigma_{\col}^{1/2}  Z_1 \big) \cdot A \Big\|_2
- \Big\|  v^\top re \big(  \Sigma_{\col}^{1/2}  Z_2 \big) \cdot A \Big\|_2 \bigg| \\
&\leq \frac{1}{\sqrt m} \bigg|  v^\top re \big(  \Sigma_{\col}^{1/2} ( Z_1 -  Z_2) \big) \cdot A  \bigg| \\
&\leq \frac{1}{\sqrt m} \| v\|_2 \Big\|  \Sigma_{\col}^{1/2} ( Z_1 -  Z_2) \Big\|_2 \cdot \|A\|_2  \\
&\leq \frac{1}{\sqrt m} \| \Sigma_{\col}^{1/2} \|_2\| Z_1 -  Z_2\|_2 \cdot \|A\|_2 \\
&= \frac{\rho_0^2}{\sqrt m} \| Z_1 -  Z_2\|_2.
\end{aligned}
\end{equation}

Using Lemma \ref{lemma:Talagrand}, we have that 
\#
\label{eq:W_concentration}
\PP\Big\{ |W - \EE W | > t \Big\} \leq 2\exp\Big(-\frac{t^2m}{2\rho_0^4}\Big).
\#
Since $W \geq 0$ and hence $\EE W \geq 0$, we have
\[
\Big[(\EE W^2)^{1/2} - \EE W \Big]^2 \leq \Big[(\EE W^2)^{1/2} + \EE W \Big] \cdot \Big[(\EE W^2)^{1/2} - \EE W \Big] = \Var(W).
\]
Moreover, from \eqref{eq:W_concentration} we have
\[
\Var(W) = \EE\Big\{ \big(W-\EE W \big)^2 \Big\} = \int_0^\infty \PP \Big\{ \big(W-\EE W \big)^2 \geq t^2 \Big\}d(t^2) \leq \int_0^\infty 2\exp\Big(-\frac{t^2m}{2\rho_0^4}\Big)d(t^2) = \frac{4\rho_0^4}{m},
\]
and hence 
\begin{equation}
\label{eq:var_bound}
(\EE W^2)^{1/2} - \EE W \leq \frac{2\rho_0^2}{\sqrt m}.
\end{equation}
According to \eqref{eq:var_bound}, we know that $|W - \EE W | \leq t $ implies $| W - (\EE W^2)^{1/2} | \leq t + 2\rho_0^2/\sqrt m$, which gives
\begin{equation}
\begin{aligned}
\PP \Big( | W - (\EE W^2)^{1/2} | > t + 2\rho_0^2/\sqrt m \Big) &\leq \PP\Big( |W - \EE W | > t \Big) \leq 2\exp\Big(-\frac{t^2m}{2\rho_0^4}\Big)
\end{aligned}
\end{equation}
for any fixed $ v \in \RR^p$ with $\| v\|_2 = 1$. 
For large enough $m$, taking $t = \frac 14 c_{\min}$ and apply union bound on 1/4-covering of $\mathbb S^{m-1} = \{ v \in \RR^m \,\, | \,\, \| v\|_2 = 1\}$ we completes the proof. 
The proof for upper bound is similar.
\end{proof}

\subsection{Proof of Lemma \ref{lemma:lambda}.}

\begin{proof}
Before we proceed with the main proof, we first introduce the following lemma in \cite{negahban2011estimation}. 
\begin{lemma}[Lemma I.2 in \cite{negahban2011estimation}]
\label{lemma:concentration_Y}
Given a Gaussian random vector $Y \sim N(0,S)$ with $Y \in \RR^{m \times 1}$, for all $t > 2/\sqrt{m}$ we have
\#
\PP\bigg[ \frac{1}{m} \Big| \|Y\|_2^2 - \tr S \Big| > 4t\|S\|_2 \bigg] \leq 2\exp\bigg( -\frac{m\big(t-\frac{2}{\sqrt m}\big)^2}{2} \bigg) + 2\exp\Big(-\frac{m}{2}\Big).
\#
\end{lemma}

We then proceed with the proof of Lemma \ref{lemma:lambda}. 
Denote $q_j =  x_j -  M^*\sum_{k\in c_j} x_k \sim N(0,  \Sigma_j)$ and denote $Q = [q_1, ..., q_m] \in \RR^{p \times m}$, we have $\EE \frac{1}{m}QQ^\top =  G$ and 
\#
\frac{1}{m} \sum_{j=1}^m\Big( x_j -  M^*\sum_{k\in c_j} x_k\Big)\cdot\sum_{k\in c_j} x_k^\top = \frac{1}{m} Q \cdot \tilde  X.
\#
For any fixed $ v \in \RR^p$ with $\| v\|_2 = 1$, we have
\begin{equation}
\begin{aligned}
\frac{1}{m}  v^\top Q \tilde  X v &= \frac{1}{m}\sum_{j=1}^m  v^\top q_j \cdot \tilde x_j^\top  v
= \frac{1}{2m} \bigg[ \sum_{j=1}^m \langle  v, q_j + \tilde x_j \rangle^2 - \sum_{j=1}^m \langle  v, q_j \rangle^2- \sum_{j=1}^m \langle  v, \tilde x_j \rangle^2 \bigg] \\
&= \underbrace{\frac{1}{2}  v^\top\bigg( \frac{1}{m} \sum_{j=1}^m (q_j + \tilde x_j)(q_j + \tilde x_j )^\top \bigg) v - \frac{1}{2}  v^\top \EE( H + QQ^\top)  v}_{R_1} \\
& \quad - \underbrace{\bigg[ \frac{1}{2}  v^\top\bigg( \frac{1}{m} \sum_{j=1}^m q_j q_j ^\top \bigg) v - \frac{1}{2}  v^\top \EE QQ^\top \cdot  v \bigg]}_{R_2}
- \underbrace{\bigg[ \frac{1}{2}  v^\top\bigg( \frac{1}{m} \sum_{j=1}^m \tilde x_j \tilde x_j ^\top \bigg) v - \frac{1}{2}  v^\top \EE H \cdot  v \bigg]}_{R_3} \\
&= R_1 - R_2 - R_3.
\end{aligned}
\end{equation}

Each $R_j$ for $j = 1, 2, 3$ is a deviation term and can be bounded similarly. For $R_3$, define the random vector $Y \in \RR^{m}$ with component $Y_j =  v^\top \tilde x_j$. Using Lemma \ref{lemma:concentration_Y} and together with assumption EC, we obtain
\#
\PP\Big[ |R_3| >  4t \sigma_{\max} \Big] \leq 2\exp\bigg( -\frac{m\big(t-\frac{2}{\sqrt m}\big)^2}{2} \bigg) + 2\exp\Big(-\frac{m}{2}\Big).
\#
Similarly, for $R_1$ and $R_2$ we have
\#
\PP\Big[ |R_2| >  4t \eta_{\max} \Big] \leq 2\exp\bigg( -\frac{m\big(t-\frac{2}{\sqrt m}\big)^2}{2} \bigg) + 2\exp\Big(-\frac{m}{2}\Big),
\#
and
\#
\PP\Big[ |R_1| >  4t (\sigma_{\max} + \eta_{\max}) \Big] \leq 2\exp\bigg( -\frac{m\big(t-\frac{2}{\sqrt m}\big)^2}{2} \bigg) + 2\exp\Big(-\frac{m}{2}\Big).
\#
Combine these three bounds, for fixed $ v \in \RR^p$ with $\| v\|_2 = 1$, we have
\#
\PP\Big[ \frac{1}{m} \Big|  v^\top Q \tilde  X v \Big|  >  8t (\sigma_{\max} + \eta_{\max}) \Big] \leq 6\exp\bigg( -\frac{m\big(t-\frac{2}{\sqrt m}\big)^2}{2} \bigg) + 6\exp\Big(-\frac{m}{2}\Big).
\#
Setting $t = 4\sqrt{p/m}$ and taking the union bound on 1/4-covering of $\mathbb S^{m-1} = \{ v \in \RR^m \,\, | \,\, \| v\|_2 = 1\}$ completes the proof.
\end{proof}

\subsection{Proof of Lemma \ref{lemma:initialization}.}

\begin{proof}
Since $ M^{(0)}$ is the unconstrained minimizer of $\cL( M)$, we have $\cL( M^{(0)}) \leq \cL( M^*)$. Since $\cL(\cdot)$ is strongly convex, we have
\[
0 \geq \cL( M^{(0)}) - \cL( M^*) \geq \langle \nabla \cL( M^*),  M^{(0)} -  M^{*} \rangle + \frac{\kappa_\mu}{2} \| M^{(0)} -  M^{*}\|_F^2.
\]
We then have
\[
\| M^{(0)} -  M^{*}\|_F^2 \leq - \frac{2}{\kappa_\mu} \langle \nabla \cL( M^*),  M^{(0)} -  M^{*} \rangle \leq \frac{2}{\kappa_\mu} \| \nabla \cL( M^*) \|_F \cdot \|  M^{(0)} -  M^{*} \|_F,
\]
and hence
\[
\| M^{(0)} -  M^{*}\|_F \leq \frac{2}{\kappa_\mu} \| \nabla \cL( M^*) \|_F \leq \frac{2\sqrt{p}\lambda}{\kappa_\mu}.
\]
For large enough $m$, this error bound can be small and Lemma 2 in \cite{yu2018recovery} gives
\#
d^2 \big( V^{(0)}, V^* \big) \leq \frac{2}{\sqrt 2-1} \cdot \frac{\| M^{(0)} -  M^{*}\|_F}{\sigma_r(M^*)} \leq\frac{20p\lambda^2}{\kappa_\mu^2 \cdot \sigma_r(M^*)}.
\#
\end{proof}

\subsection{Proof of Theorem \ref{theorem:nonconvex}.}

\begin{proof}
According to Lemma \ref{lemma:lambda} and Lemma \ref{lemma:initialization}, the initialization $ M^{(0)}$ satisfies $\| M^{(0)} -  M^{*}\|_F \leq C$ as long as $m \geq 4C_0p^2/\kappa_{\mu}^2$. 
Furthermore, Lemma \ref{lemma:sigma_min(H)} shows that the objective function $\cL(\cdot)$ is strongly convex and smooth. 
%Theorem \ref{theorem:nonconvex} then follows from Theorem 1 in \cite{yu2018recovery}. 
Therefore we apply Lemma 3 in \cite{yu2018recovery} and obtain
\begin{equation}
\label{eq:lemma_contraction}
d^2 \Big( V^{(t+1)} , V^* \Big) \leq  \Big(1 - \eta\cdot
\frac{2}{5} \mu_{\min}\sigma_{M}
\Big)\cdot d^2 \Big( V^{(t)}, V^* \Big)
 +
\eta\cdot\frac{\kappa_{L}  + \kappa_{\mu} }{\kappa_{L} \cdot\kappa_{\mu} }\cdot e_{\rm stat}^2 ,
\end{equation}
where $\mu_{\min} = \frac 18 \frac{\kappa_\mu\kappa_L}{\kappa_\mu + \kappa_L}$ and $\sigma_{M} = \|M^*\|_2$. Define the contraction value
\begin{equation}
\label{eq:def_beta}
\beta = 1 - \eta\cdot
\frac{2}{5} \mu_{\min}\sigma_{M} < 1, 
\end{equation}
we can iteratively apply \eqref{eq:lemma_contraction} for each $t = 1, 2, ..., T$ and obtain
\begin{equation}
d^2 \Big( V^{(T)}, V^* \Big) \leq \beta^T d^2 \Big( V^{(0)}, V^* \Big) +  \frac{\eta}{1-\beta}\cdot\frac{\kappa_{L}  + \kappa_{\mu} }{\kappa_{L} \cdot\kappa_{\mu}}\cdot e_{{\rm stat}}^2,
\end{equation}
which shows linear convergence up to statistical error. For large enough $T$, the final error is given by
\begin{equation}
\begin{aligned}
\frac{\eta}{1-\beta}\cdot\frac{\kappa_{L}  + \kappa_{\mu} }{\kappa_{L} \cdot\kappa_{\mu}}\cdot e_{{\rm stat}}^2 
&= \frac{5}{2\mu_{\min} \sigma_M} \cdot\frac{\kappa_{L}  + \kappa_{\mu}}{\kappa_{L} \cdot\kappa_{\mu}}\cdot e_{{\rm stat}}^2 \\
&= \frac{20}{\sigma_M} \cdot \Big( \frac{\kappa_{L}  + \kappa_{\mu}}{\kappa_{L} \cdot\kappa_{\mu}} \Big) ^2\cdot e_{{\rm stat}}^2 \\
& \leq \frac{80}{\sigma_M} \cdot \frac{e_{{\rm stat}}^2}{\kappa^2_\mu}.
\end{aligned}
\end{equation}
Together with \eqref{eq:stat_error_rate} we see that this gives exactly the same rate as the convex relaxation method~\eqref{eq:convex_rate}.
\end{proof}

\end{document}